\documentclass[10pt,twocolumn,letterpaper]{article}

\usepackage{cvpr}
\usepackage{times}
\usepackage{epsfig}
\usepackage{graphicx}
\usepackage{amsmath}
\usepackage{amssymb}
\usepackage{booktabs}
\usepackage{multirow}
\usepackage{subfig}
\usepackage{url}

\usepackage[pagebackref=true,breaklinks=true,letterpaper=true,colorlinks,bookmarks=false]{hyperref}

\cvprfinalcopy 

\newcommand{\tabincell}[2]{\begin{tabular}{@{}#1@{}}#2\end{tabular}} 

\makeatletter
\DeclareRobustCommand\onedot{\futurelet\@let@token\@onedot}
\def\@onedot{\ifx\@let@token.\else.\null\fi\xspace}

\def\ie{\emph{i.e}\onedot}

\def\etal{\emph{et al}\onedot}

\def\cvprPaperID{3} 

\ifcvprfinal\fi
\begin{document}

\title{Reducing the feature divergence of RGB and near-infrared images using Switchable Normalization}

\author{Siwei Yang\thanks{Equal contribution to this work}, \quad Shaozuo Yu\footnotemark[1], \quad Bingchen Zhao\footnotemark[1], \quad Yin Wang\\
Tongji University, Shanghai, China\\
{\tt\small \{billyang426, yushaozuo, zhaobc, yinw\}@tongji.edu.cn}
}

\maketitle
\thispagestyle{empty}

\begin{abstract}
    Visual pattern recognition over agricultural areas is an important application of aerial image processing.
    In this paper, we consider the multi-modality nature of agricultural aerial images and show that naively combining different modalities together without taking the feature divergence into account can lead to sub-optimal results.
    Thus, we apply a Switchable Normalization block to our DeepLabV3+ segmentation model to alleviate the feature divergence.
    Using the popular symmetric Kullback–Leibler divergence measure, we show that our model can greatly reduce the divergence between RGB and near-infrared channels.
    Together with a hybrid loss function, our model achieves nearly 10\% improvements in mean IoU over previously published baseline.
\end{abstract}

\section{Introduction}
\label{sec:intro}

Recent progress in CNNs demonstrates significant improvements in many typical computer vision tasks such as classification, object detection and segmentation~\cite{krizhevsky2012imagenet,deng2009imagenet,Chen2016DeepLabSI,chen2018deeplabv3+}.
Applications in multiple domains developed rapidly by employing deep neural networks and achieved better accuracy and efficiency. 

Agriculture, as one of the most fundamental fields for humanity, is a significant application area of computer vision. 
One important direction of visual pattern recognition in agriculture is aerial image semantic segmentation.
Different from conventional image semantic segmentation dataset where only RGB based image is available~\cite{Cordts2016Cityscapes,zhou2017scene,zhou2016semantic,lin2014microsoft}, the agricultural data collection process utilizes specific cameras to capture Red, Green and Blue channel(RGB) with an additional near-infrared(NIR) signal channel which can be used in the pattern recognition process~\cite{chiu2020agriculture}.
Also, agricultural data is naturally imbalanced.

\begin{figure}[ht]
    \centering
    \includegraphics[width=1\linewidth]{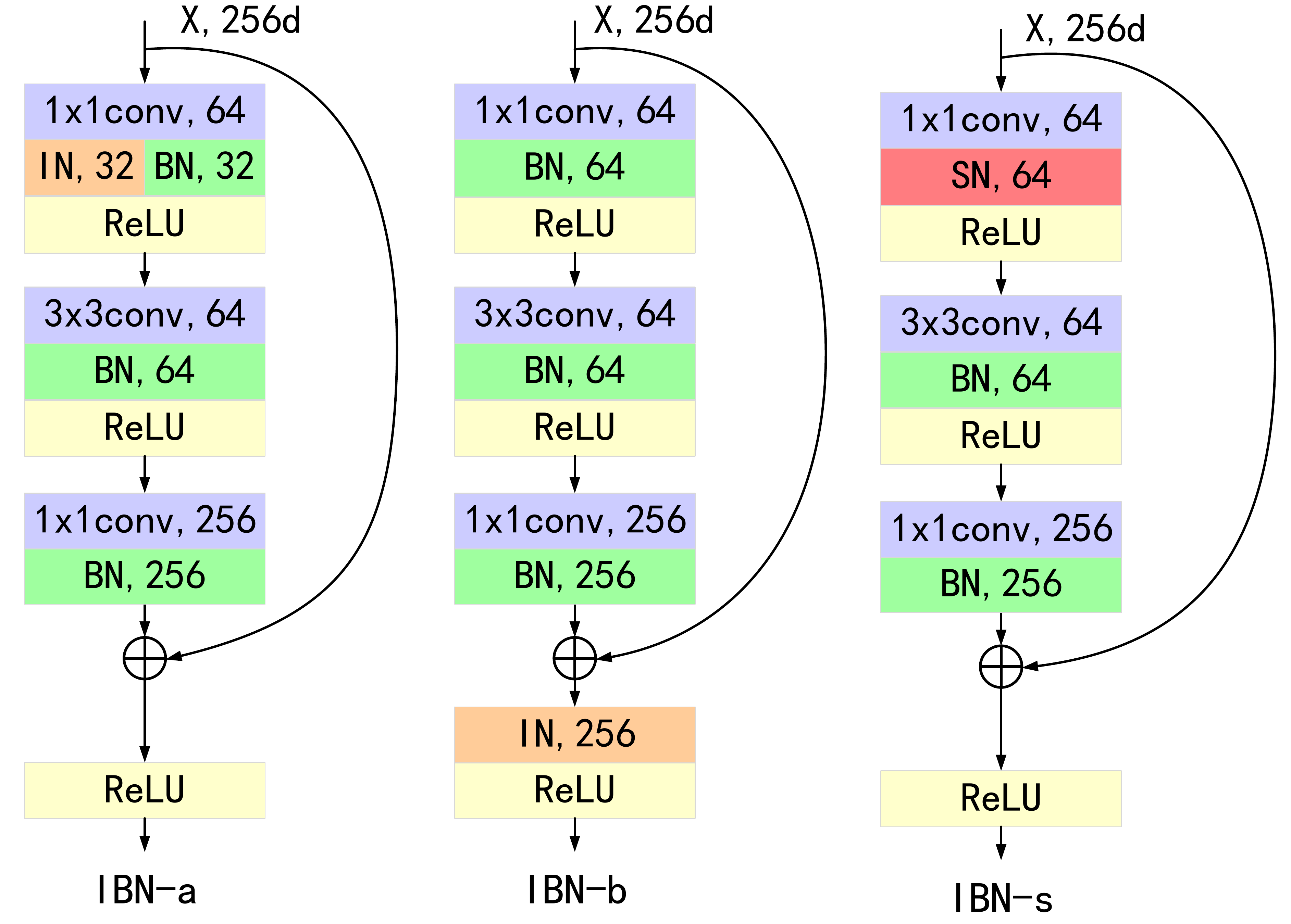}
    \caption{Instance-batch normalization (IBN) block. IBN-a and IBN-b are the variants of IBN-Net proposed in~\cite{pan2018IBN-Net}, we change the first BN to Switchable-Normlization for better performance.}
    \label{fig:IBN-stucture}
\end{figure}

In~\cite{chiu2020agriculture}, the authors proposed to add the additional NIR channel to the corresponding RGB image to form a Near-infrared-Red-Green-Blue(NRGB) 4-channel image.
By duplicating the weights corresponding to the Red channel of the first convolution layer, the authors can utilize imagenet pretrained model for NRGB images.
In their experiment, the model utilizes the NRGB images for training and testing gains 2.92\% improvement in mean intersection-over-union (mIoU) over the RGB image count part.
This demonstrated that using NRGB images is more effective than using RGB images only.
However, by employing the technique called symmetric KL divergence which is used in AdaBN~\cite{li2016revisiting}, we found that there is a feature divergence between RGB and NIR images.
This feature divergence between RGB and NIR images can be seen as the representation of the inherent data modality or domain difference between RGB and NIR images.
Naively using the NRGB images without taking the feature divergence into account can lead to sub-optimal results.

\begin{figure*}[ht]
\begin{center}
\centering
\includegraphics[width=1.0\linewidth]{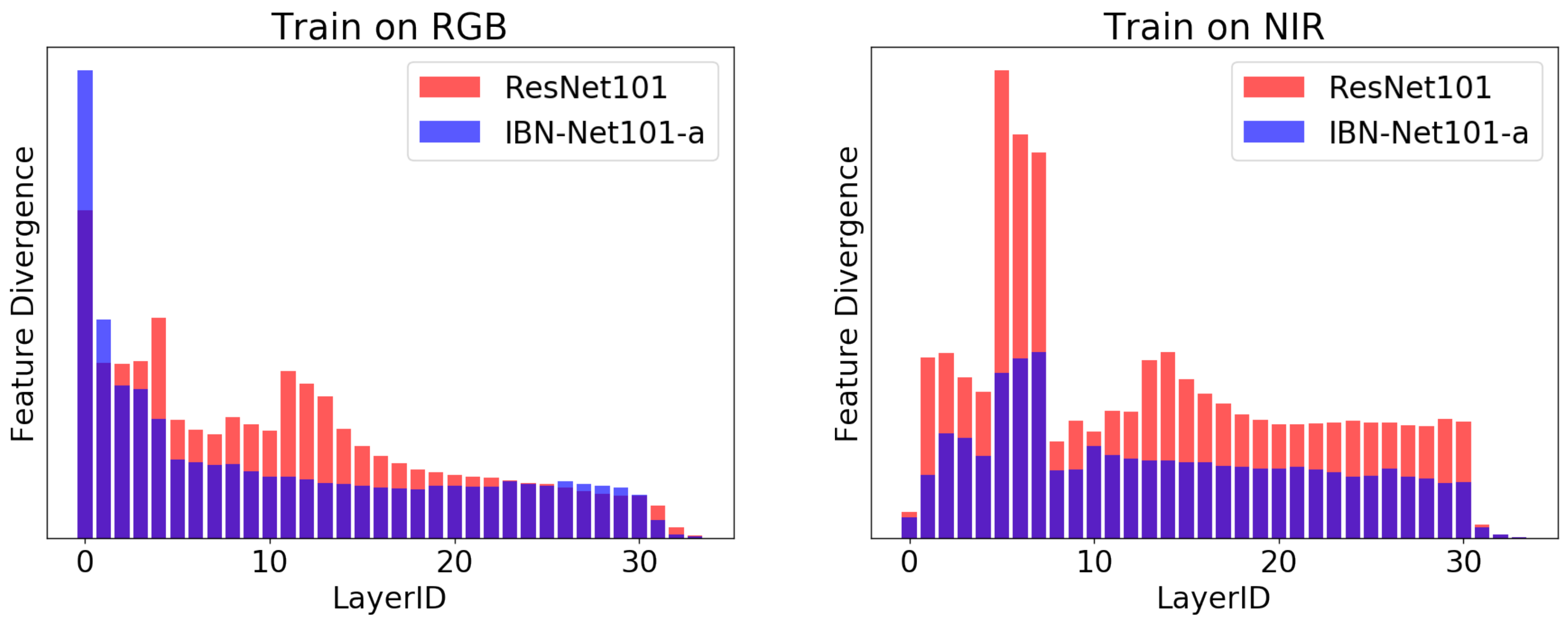}
  \caption{Both figures represent feature divergence between RGB images and NIR images. The vertical axis indicates the feature divergence between RGB images and NIR images. IBN-Net101-a reduces the feature divergence dramatically when trained on RGB images or NIR images.}
\label{fig:Feature Divergence}
\vspace{-0.2cm}
\end{center}
\end{figure*}

In our experiments, we found that the recently proposed IBN-Net~\cite{pan2018IBN-Net} can reduce the feature divergence between two data modalities effectively.
The IBN-Net incorporates Instance Normalization(IN)~\cite{ulyanov2016instance} and Batch Normalization~\cite{ioffe2015batch} together to enhance the generalization capacity on different domains.
In Fig.~\ref{fig:Feature Divergence}, we show the feature divergence between RGB and NIR can be reduced by using IBN-Net.

By further modify the original IBN-Net architecture (IBN-s) and using a hybrid loss function which addresses the imbalanced data problem and directly optimize the evaluation metric, we achieve a 4.03\% improvement in mIoU on the validation set of Agriculture-Vision challenge dataset and scored \textbf{54.0\%} mIoU on the test set.

To summarize, first we show that the feature divergence between RGB and NIR images should be carefully addressed to achieve better performance on the agricultural data. 
Second, we propose to use a widely adopted method called symmetric KL divergence to measure the feature divergence between RGB and NIR images.
Finally, based on the motivation of reducing the feature divergence between RGB and NIR images, we proposed a novel network building block called IBN-s, and achieves comparable results on both the validation and test set of the Agriculture-Vision Challenge dataset.

\section{Feature Divergence Analysis}
\label{sec:feature divergence}

Appearance difference between RGB and NIR images can cause huge feature divergence which results in bad cross-modality generalization.

\subsection{Direct generalization of two modalities}
\label{subsec:direct gen}

Here we present the experimental result of training a DeepLabV3+~\cite{chen2018deeplabv3+} model with ResNet101~\cite{he2016deep} as the backbone on one data modalities and test on the other, \ie, train on RGB and test on NIR.

\begin{table}[]
\begin{center}
\begin{tabular}{ccc}
\toprule
\multirow{3}{*}{} & \multicolumn{2}{c}{mIoUs(\%)} \\
\cmidrule(r){2-3}
& Train on RGB & Train on NIR\\
\midrule
Test on RGB & 46.05  &  23.39 \\
Test on NIR & 18.29  &  44.22 \\
Perf. Decay &  27.76   &  20.83   \\
\bottomrule
\end{tabular}
\end{center}
\caption{`Perf. Decay' stands for ``Performance decay''. When models are trained on NIR images alone, NIR images with a single channel will be duplicated channel-wise to transform into a three-channel images. All the other settings about the experiment regarding this table are the same as ones in Section~\ref{sec:experiment}. 
Testing on different modalities from training can reduce the performance by a huge margin. This phenomenon suggests that the feature divergence between RGB and NIR can be huge.}
\label{tab:generalization}
\vspace{-0.1cm}
\end{table}

\begin{figure*}[htbp]
\centering
\subfloat[cloud shadow]
{
    \includegraphics[width=0.16\linewidth]{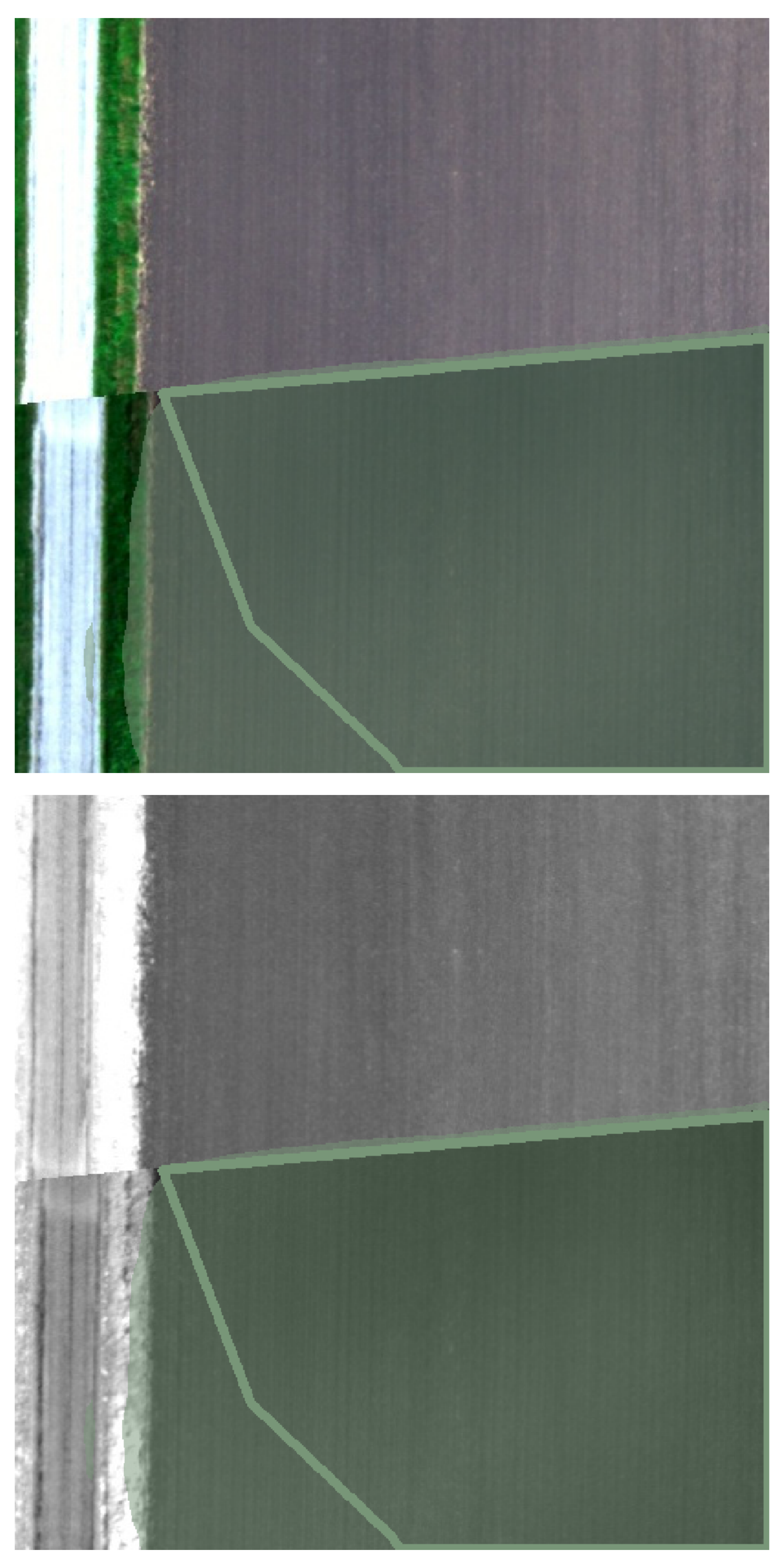}
}
\subfloat[double plant]
{
    \includegraphics[width=0.16\linewidth]{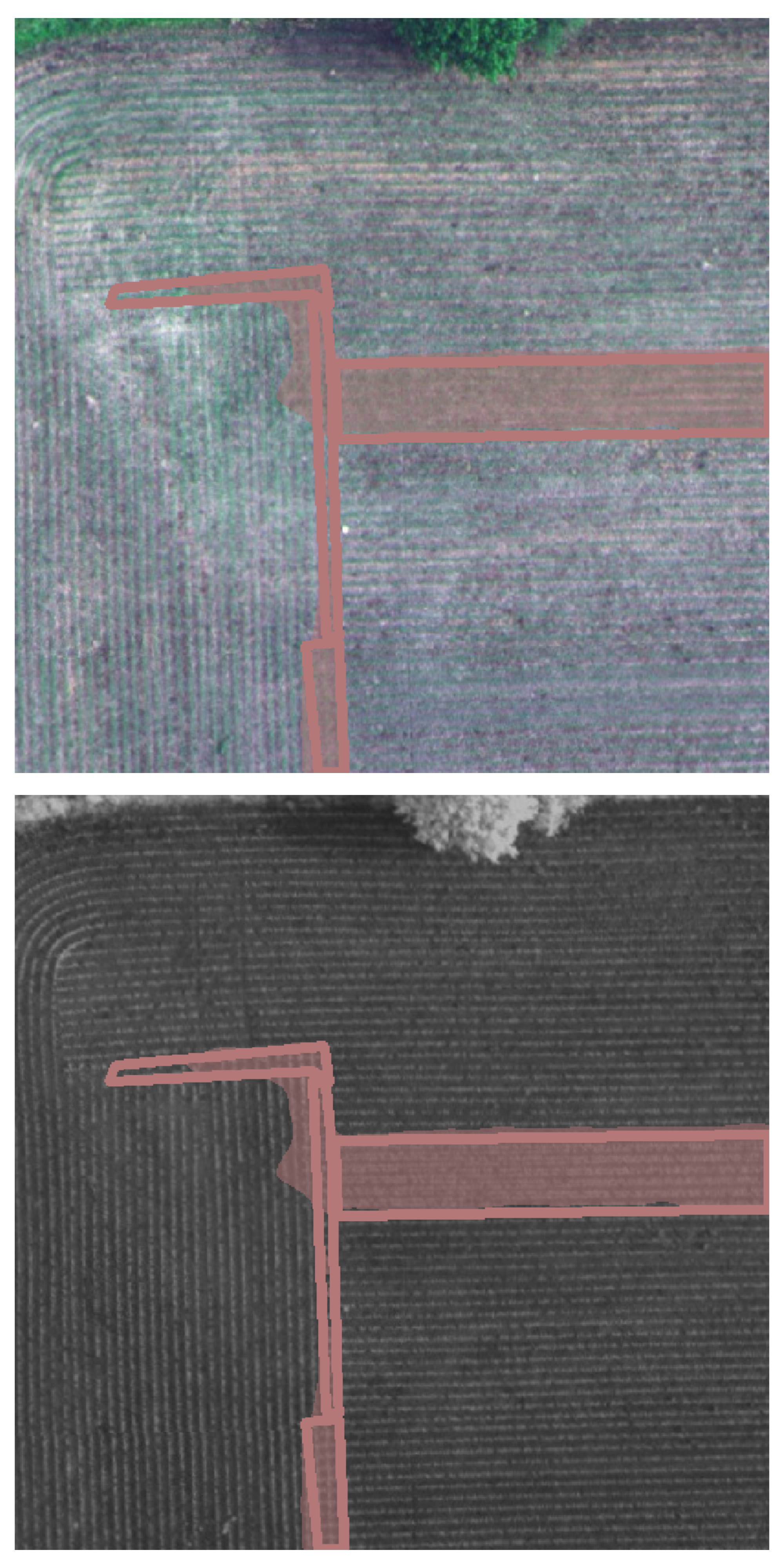}
}
\subfloat[planter skip]
{
    \includegraphics[width=0.16\linewidth]{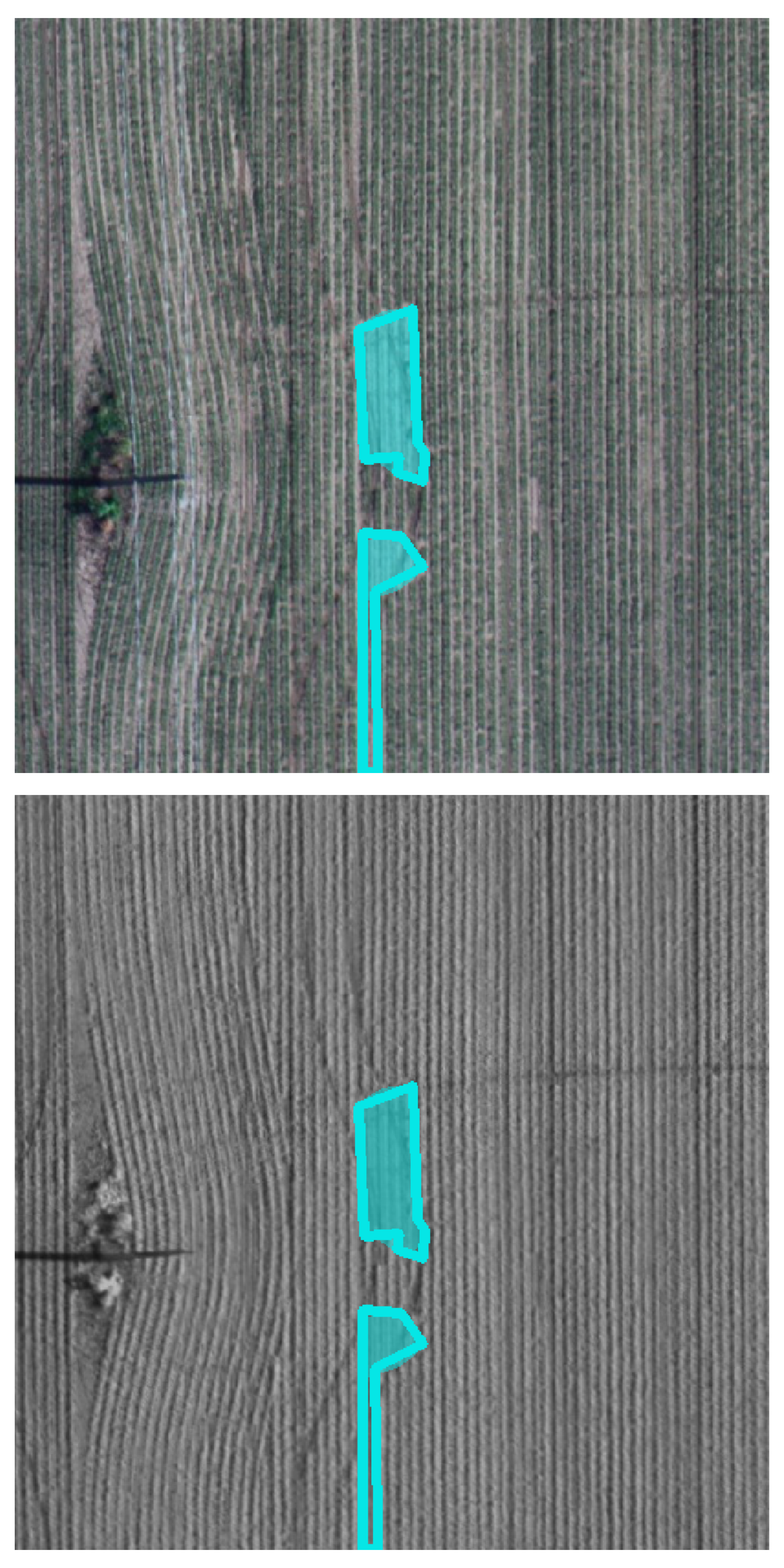}
}
\subfloat[standing water]
{
    \includegraphics[width=0.16\linewidth]{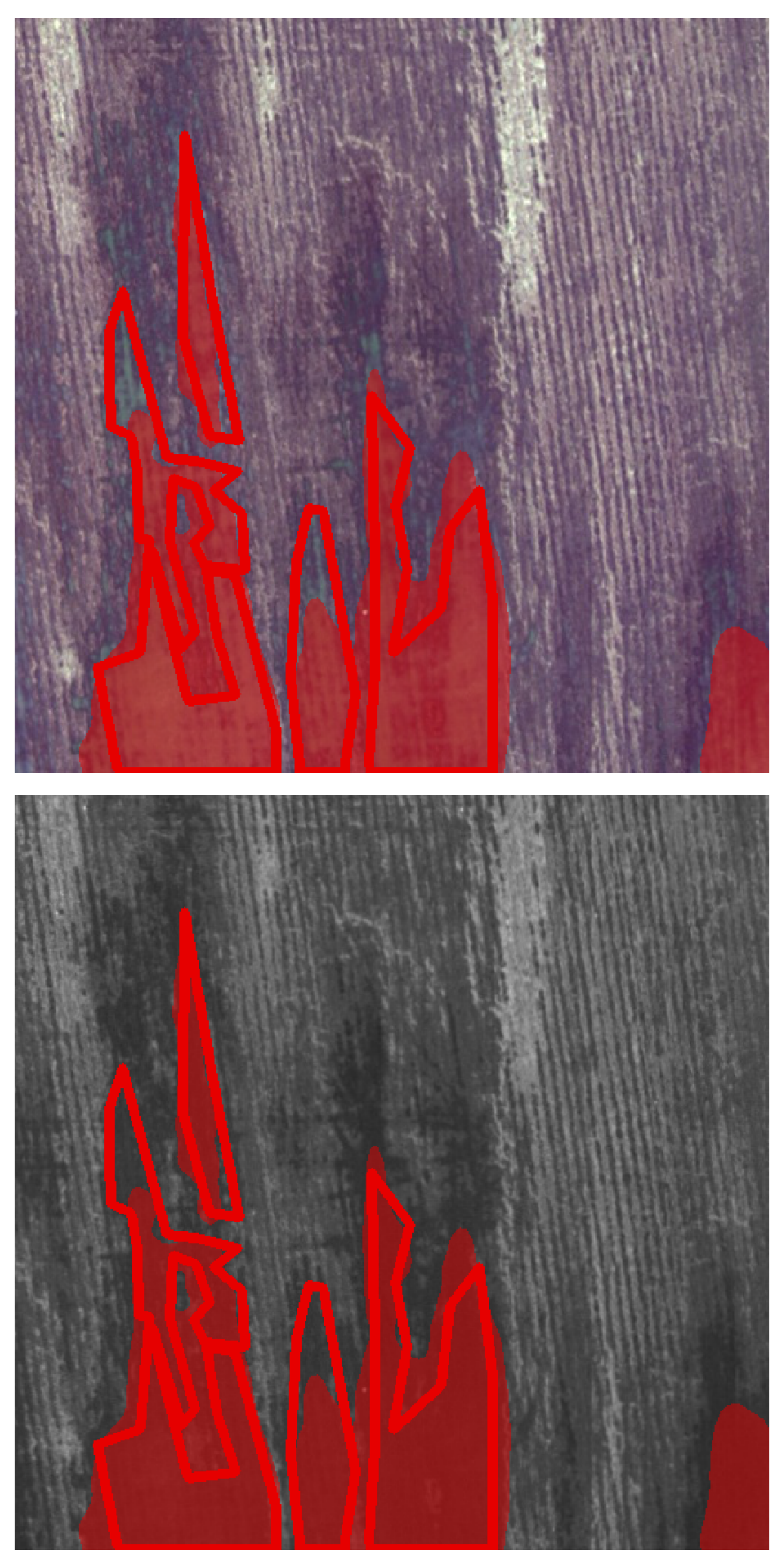}
}
\subfloat[waterway]
{
    \includegraphics[width=0.16\linewidth]{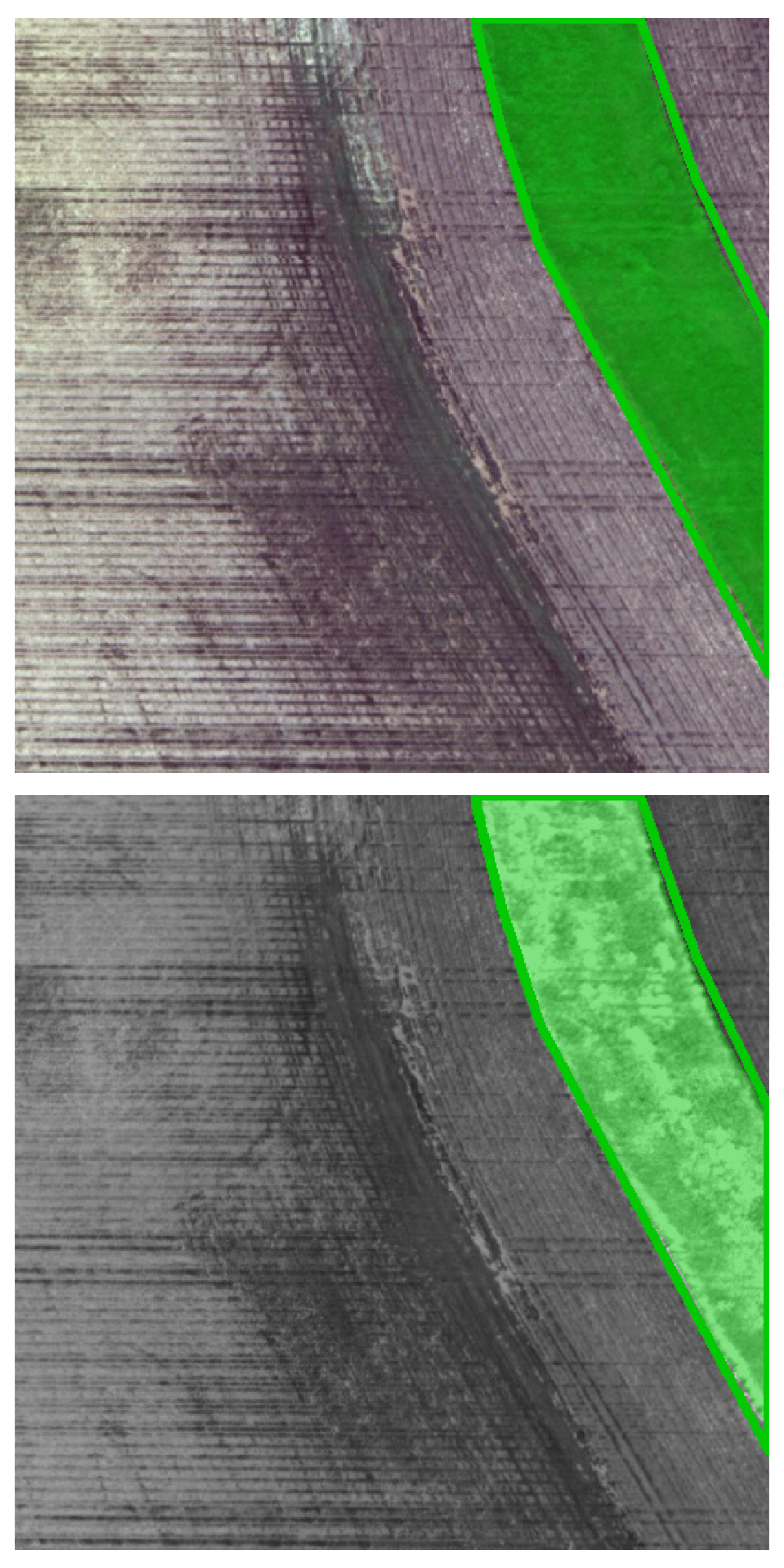}
}
\subfloat[weed cluster]
{
    \includegraphics[width=0.16\linewidth]{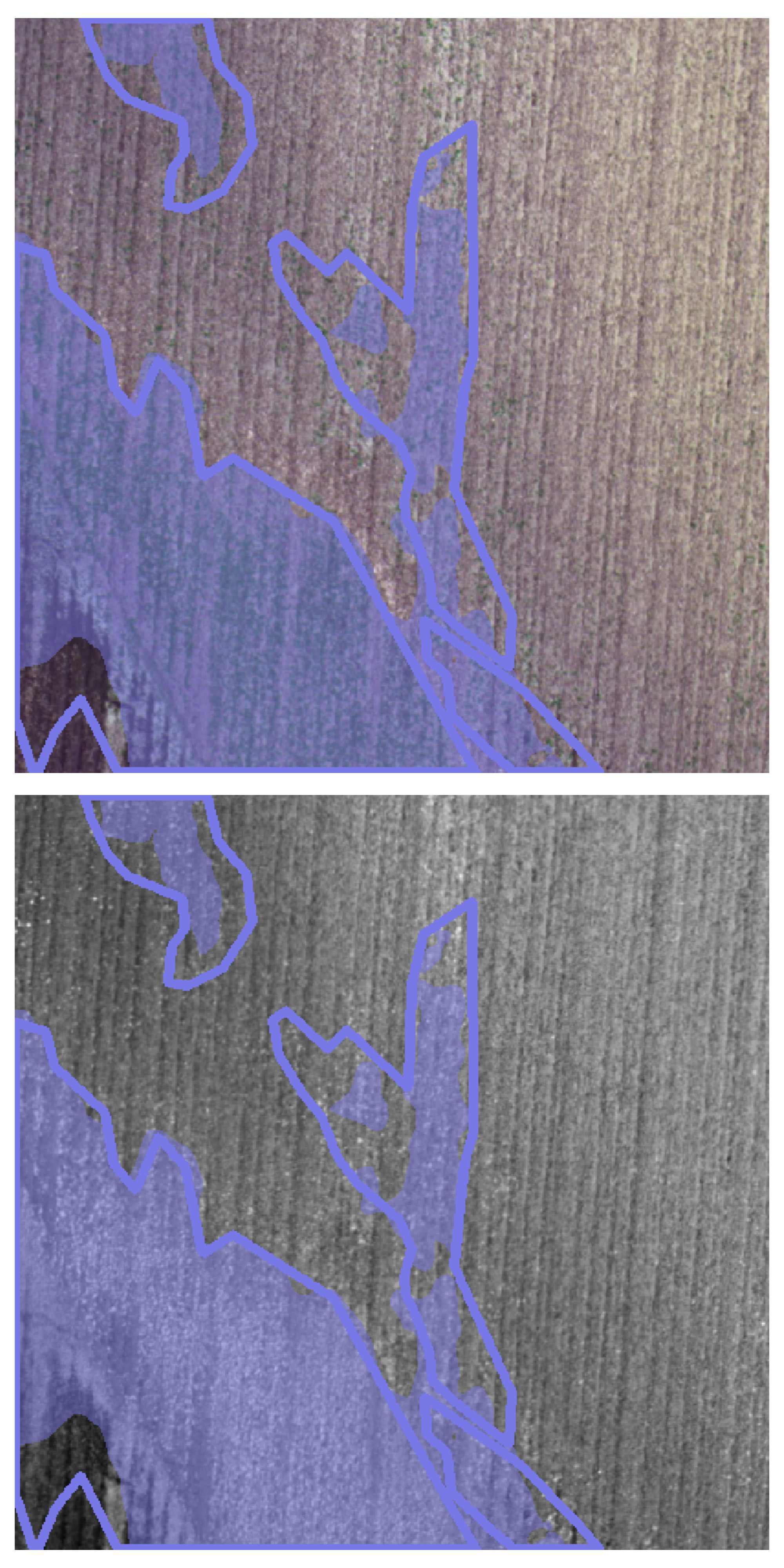}
}
\caption{For each class, top: RGB image; bottom: NIR image; line: ground truth; mask: prediction. All predictions are given by IBN-Net101-s trained with BCE+Dice+Lovasz as loss function.}
\label{fig:example}
\end{figure*}

\subsection{Symmetric KL Divergence}
\label{subsec:KLD}
In order to quantize the feature divergence between two data modalities, we denote each feature $i$ as $F_i$, and assume that the feature follows a Gaussian distribution with mean $\mu_i$ and variance $\sigma^2_i$.
Then we can naturally use a symmetric KL divergence as our metric for the feature divergence between data modalities:

\begin{align}
    KL(A\|B)&=log\frac{\sigma_{A}}{\sigma _{B}}+\frac{\sigma_{A}^{2}+(\mu_{A}-\mu_{B})^{2}}{2\sigma_{B}^{2}}-\frac{1}{2}\label{equ:KL_div}\\
    D(A\|B)&=KL(A\|B)+KL(B\|A)\label{equ:sym_KLD}
\end{align}

\subsection{Feature Divergence}
\label{subsec:feature divergence}
Symmetric KL divergence can be used to measure the distance between feature distribution in different modalities.

For a certain layer in one model, we calculate the mean and variance on each channel among every sample in the validation set. The mean symmetric KL divergence of every channel of a layer is defined as the feature divergence of this layer.
Thus, for a layer with C channels, this layer's feature divergence with features $F_A$ and $F_B$ respectively on two modalities are defined as below:

\begin{align}
    D(L_A\|L_B) = \frac{1}{C} \sum_{i=1}^{C}D(F_{iA} \| F_{iB}) \label{equ:feature_div}
\end{align}

To understand how IBN-Net achieves better performance on NRGB images, we compared the feature divergence of 34 ReLU layers in both IBN-Net101-a and ResNet101 on RGB and NIR images.

As Figure~\ref{fig:Feature Divergence} shown, the feature divergence caused by different appearances in RGB and NIR images are obviously reduced among most layers in IBN-Net101.

\begin{table*}[htb]
\begin{center}
\begin{tabular}{ccccccccc}
\toprule
\multirow{3}{*}{Backbone} & \multicolumn{8}{c}{IoUs(\%)} \\
\cmidrule(r){2-9} 
& Background & \tabincell{c}{Cloud\\shadow} & \tabincell{c}{Double\\plant} & \tabincell{c}{Planter\\skip} & \tabincell{c}{Standing\\Water} & Waterway & \tabincell{c}{Weed\\Cluster} & \tabincell{c}{Mean\\IoU} \\
\midrule
ResNet50 & 79.95 & 45.75 & 36.98 & 1.18 & 59.67 & 58.03 & 48.58 & 47.16 \\
IBN-Net50-a & 80.51 & \textbf{53.67} & 40.87 & 1.83 & 64.44 & \textbf{61.67} & 49.71 & 50.39 \\
IBN-Net50-b & 80.41 & 52.48 & 39.32 & \textbf{4.19} & 62.82 & 57.91 & 49.36 & 49.50 \\
IBN-Net50-s & \textbf{80.82} & 53.52 & \textbf{43.63} & 3.59 & \textbf{65.44} & 60.20 & \textbf{50.84} & \textbf{51.15} \\
\midrule
ResNet101 & 79.32 & 47.95 & 39.77 & 0.98 & 62.47 & 61.17 & 49.36 & 48.72 \\
IBN-Net101-a & 80.79 & 52.64 & 38.27 & 2.72 & \textbf{67.52} & 61.96 & 48.52 & 50.35 \\
IBN-Net101-b & \textbf{80.88} & 52.05 & 40.75 & 3.19 & 64.21 & 59.88 & \textbf{51.05} & 50.29 \\
IBN-Net101-s & 80.78 & \textbf{52.69} & \textbf{44.53} & \textbf{3.34} & 66.26 & \textbf{62.26} & 50.39 & \textbf{51.46} \\
\bottomrule
\end{tabular}
\end{center}
\caption{Different backbones setting with DeepLabV3+ architecture and BCE+Dice as loss function on NRGB images. The detailed loss defintion is given in~\ref{sec:loss}. IBN-Net101-s achieves highest mIoU of 51.46\% on val set.}
\label{tab:backbone}
\end{table*}

\subsection{Discussion}
\label{subsec:dicussion}
It is shown that IBN-Net reduces the distance between feature distributions in RGB and NIR images.
The analysis on feature divergence gives us an intuition of how IBN-Net gains stronger performance on NRGB images since regular ResNet needs to learn to adapt appearance variance between RGB and NIR images while IBN-Net can automatically filter out this variance by introducing IN into the model.

\section{Our Model}

Our model is based on Deeplabv3+~\cite{chen2018deeplabv3+} with IBN-Net as the backbone.

\subsection{IBN-Net}
Instance Normalization~\cite{ulyanov2016instance}(IN) learns appearance-related features, while Batch Normalization~\cite{ioffe2015batch}(BN) preserves content-related information.
By integrating IN and BN, IBN-Net achieves better generalization and performance on various computer vision tasks~\cite{pan2018IBN-Net}.

Pan~\etal~\cite{pan2018IBN-Net} introduced two kinds of IBN-block, IBN-a and IBN-b.
These two kinds of IBN block work differently.
IBN-a introduces IN into the residual path while IBN-b introduces IN into the identity path(Figure~\ref{fig:IBN-stucture}).

As shown in Table~\ref{tab:backbone} and Figure~\ref{fig:Feature Divergence}, IBN improves mIoU on NRGB images by reducing feature divergence between RGB and NIR images.

\subsection{IBN-s}
Figure~\ref{fig:Feature Divergence} shows that the feature divergence is reduced in deeper layers.
The effect of IN removing appearance variant features is weakened while the effect of BN preserving discriminative information is enhanced as the depth of layers grows.
However, the integration method between IN and BN is fixed among all the layers, unable to adjust according to depth.

Thus, we propose IBN-s by replacing concatenated IN and BN in IBN-a with Switchable Normalization(SN)~\cite{luo2018differentiable}.
With IBN-s, now models can learn how to trade off IN and BN's effect at different depths.

As shown in Table~\ref{tab:backbone}, our IBN-s improves mIoU by 2.74\% compared with ordinary ResNet Bottleneck, and 1.11\% compared with IBN-a.

\begin{table*}[htb]
\begin{center}
\begin{tabular}{cccccccccc}
\toprule

\multirow{3}{*}{$\lambda_1$} & \multirow{3}{*}{$\lambda_2$} & \multicolumn{8}{c}{IoUs(\%)} \\
\cmidrule(r){3-10} 
& & Background & \tabincell{c}{Cloud\\shadow} & \tabincell{c}{Double\\plant} & \tabincell{c}{Planter\\skip} & \tabincell{c}{Standing\\Water} & Waterway & \tabincell{c}{Weed\\Cluster} & \tabincell{c}{Mean\\IoU} \\
\midrule
0 & 0 & 80.49     & 51.09     & 40.98     & 3.77     & 63.75     & 57.01     & 49.75     & 49.55 \\
1 & 0 & 80.79 & 52.64 & 38.27 & 2.72 & \textbf{67.52} & 61.96 & 48.52 & 50.35 \\
0 & 1 & \textbf{81.35} & 55.43    & 40.33     & 1.42     & 65.70     & \textbf{64.30}     & 48.83     & 51.05 \\
1 & 1 & 81.06 & \textbf{57.56} & \textbf{46.30} & \textbf{12.45}     & 65.11     & 60.63     & \textbf{51.55}     & \textbf{53.52} \\
\bottomrule
\end{tabular}
\end{center}
\caption{Different loss settings with IBN-Net101-a+DeepLabV3+ on NRGB images. ($\lambda_1, \lambda_2$)=(1, 1) achieves highest mIoU of 53.52\% on val set.}
\label{tab:different losses}
\end{table*}

\section{Hybrid Loss}
\label{sec:loss}

With our severely imbalanced dataset, the regular cross-entropy loss~\ref{equ:bce_loss} results in slow convergence and bad performance.

Dice loss~\ref{equ:dice_loss}~\cite{milletari2016v} is a commonly used loss function in semantic segmentation tasks.
It can directly improve IoU while has a smoother gradient than using negative IoU as loss function.
Unlike pixel-wise cross-entropy loss, dice loss automatically ignores the negative pixels in the label.
Thus, it leads to a better performance than cross-entropy loss on imbalanced data.

\begin{align}
    L_{BCE}&=-\frac{1}{N}\sum_{i=1}^N{{y_i\log{\sigma(s_i)} + (1 - y_i)\log{(1 - \sigma(s_i))}}}\label{equ:bce_loss}\\
    L_{Dice}&=-\frac{1}{N}\sum_{i=1}^N{\frac{2y_i\sigma(s_i)}{y_i + \sigma(s_i)}}\label{equ:dice_loss}
\end{align}

Since there is an overlap between classes in the agricultural dataset, we consider this $m$-class segmentation task as $m$ independent binary segmentation tasks.
Lovász hinge loss~\cite{berman2018lovasz} is another attempt to directly optimizes IoU for binary segmentation tasks. 
It is a pixel-wise convex surrogate to the IoU loss based on the Lovász extension of submodular set functions.
It often yields better performance compared to the IoU loss.  

Define $m \in \mathbb{R}^p$ as $m_i = 1 - sign(y_i)*s_i$ with

\begin{equation}
sign(x) =
\begin{cases}
1 &x>0\\
-1 &x\le 0
\end{cases}
\end{equation}

Define $\hat{m} \in \mathbb{R}^p$ as $\hat{m}_i = m_{\pi_i}$ as $\pi$ being a permutation ordering the components of $m$ in a decreasing order, $\ie{m_{\pi_1} \ge m_{\pi_2}\dots\ge m_{\pi_p}}$

Define $\hat{y} \in \mathbb{R}^p$ as $\hat{y}_i = y_{\pi_i}$ and $s$ is cumulative sum of elements of $\hat{y}, \ie{s_i = \sum_{j=1}^i\hat{y}_j}$

\begin{align}
    \Delta_i &= 1 - \frac{\sum\limits_{j=i+1}^p{\hat{m}_j}}{\sum\limits_{j=1}^p{\hat{m}_j} + \sum\limits_{j=1}^i{(1 - \hat{m}_j)}}\\
    \hat{\Delta}_i &=
    \begin{cases}
    \Delta_i &i=1\\
    \Delta_i-\Delta_{i-1} &i>1
    \end{cases}\\
    L_{Lovasz} &= \sum_{i=1}^p{ReLU(m_i)*\hat{\Delta}_i} 
\end{align}




To summarize these loss functions, we purpose a hybrid loss composed of binary cross-entropy loss, dice loss and Lovasz loss defined as \ref{equ:hybrid_loss}.

\begin{align}
    L_{hybrid}&=\frac{1}{1 + \lambda_1 + \lambda_2} (L_{BCE} + \lambda_1 L_{Dice} + \lambda_2 L_{Lovasz})\label{equ:hybrid_loss}
\end{align}

Here $y$ and $y'$ stand for target and logits vector for all the pixels in one image.
$\lambda_1$ and $\lambda_2$ controls the trade-off between different losses.

As shown in Table~\ref{tab:different losses}, Our hybrid loss achieves the highest mIoU of 53.52 among all the loss settings we tried.

\section{Experiment Details}
\label{sec:experiment}

Our experiments are performed on the Agriculture-Vision challenge dataset~\cite{chiu2020agriculture}~\footnote{Our code is publicly available at \url{https://github.com/LAOS-Y/AgriVision}}.
We use DeepLabv3+~\cite{chen2018deeplabv3+} with various backbone~\cite{he2016deep, pan2018IBN-Net, luo2018differentiable} in our experiments.
All backbones are pretrained on ImageNet.
We train each model for 25,000 iterations with a batch size of 32.
We use SGD with momentum as the optimizer~\cite{sutskever2013Momentum}.
Momentum and weight decay are set to 0.9, 5e-4 respectively.
First we warmup the learning rate for 1,000 iterations~\cite{goyal2017accurate}, then train the model for 7,000 iterations with a constant learning rate of 0.01.
We then apply the `poly' learning rate policy with the power of 0.9 in the remaining 17,000 iterations.

\section{Conclusion}
\label{sec:conclusion}

In this paper, we show that despite using NIR and RGB images together to perform the agricultural pattern recognition tasks can improve the baseline performance.
There is still a feature divergence between NIR and RGB data modalities.
Our experiments show that IBN-Net~\cite{pan2018IBN-Net} and its variants can improve the model performance by reducing the divergence between features.
Inspired by IBN, we designed a novel IBN-s block to better tackle this feature divergence problem, combined with our hybrid loss function, our IBN-s model achieves over 10\% improvement over the baseline result proposed by Chiu~\etal~\cite{chiu2020agriculture}.
This suggests that reducing the feature divergence between different data modalities can be a promising direction to further improves the performance on Agricultural pattern recognition tasks.

\section{Future Work}
\label{sec:future Work}

The motivation of our work is to reduce the feature divergence between RGB and NIR images, the rationale behind this idea is the fact that there indeed is an appearance variance between RGB and NIR modalities.
In our experiments, by introducing Instance Norm~(IN) which can filter out the appearance variance into the model, IBN-Net~\cite{pan2018IBN-Net} achieves better mIoU score on both validation and test set.
However, our work has not included any explicit proof that IBN-Nets lead to a better mIoU score by lowering feature divergence.
Further analysis about how IBN-Nets improve the performance of the model is essential.
Our future work may include using style-transfer~\cite{huang2017adain} or CycleGAN~\cite{CycleGAN2017} models to close the appearance gap between two modalities, and then test the effectiveness of IBN-Net~\cite{pan2018IBN-Net} and our modified IBN-s Net on the style-transferred dataset to further validate that the gain of mIoU comes from the ability of removing the appearance variance between modalities rather than the increased model capacity of IBN-Net.
Also, if we confirmed that the performance gain indeed is from the ability of removing the appearance variance, we plan to find theoretical proof to support our findings using learning theory.

{\small
\bibliographystyle{ieee_fullname}
\bibliography{egbib}

\begin{thebibliography}{10}\itemsep=-1pt

\bibitem{berman2018lovasz}
Maxim Berman, Amal Rannen~Triki, and Matthew~B Blaschko.
\newblock The lov{\'a}sz-softmax loss: a tractable surrogate for the
  optimization of the intersection-over-union measure in neural networks.
\newblock In {\em Proceedings of the IEEE Conference on Computer Vision and
  Pattern Recognition}, pages 4413--4421, 2018.

\bibitem{Chen2016DeepLabSI}
Liang-Chieh Chen, George Papandreou, Iasonas Kokkinos, Kevin Murphy, and
  Alan~L. Yuille.
\newblock Deeplab: Semantic image segmentation with deep convolutional nets,
  atrous convolution, and fully connected crfs.
\newblock volume~40, pages 834--848, 2016.

\bibitem{chen2018deeplabv3+}
Liang-Chieh Chen, Yukun Zhu, George Papandreou, Florian Schroff, and Hartwig
  Adam.
\newblock Encoder-decoder with atrous separable convolution for semantic image
  segmentation.
\newblock In {\em ECCV (7)}, volume 11211, pages 833--851. Springer, 2018.

\bibitem{chiu2020agriculture}
Mang~Tik Chiu, Xingqian Xu, Yunchao Wei, Zilong Huang, Alexander Schwing,
  Robert Brunner, Hrant Khachatrian, Hovnatan Karapetyan, Ivan Dozier, Greg
  Rose, et~al.
\newblock Agriculture-vision: A large aerial image database for agricultural
  pattern analysis.
\newblock {\em arXiv preprint arXiv:2001.01306}, 2020.

\bibitem{Cordts2016Cityscapes}
Marius Cordts, Mohamed Omran, Sebastian Ramos, Timo Rehfeld, Markus Enzweiler,
  Rodrigo Benenson, Uwe Franke, Stefan Roth, and Bernt Schiele.
\newblock The cityscapes dataset for semantic urban scene understanding.
\newblock In {\em Proc. of the IEEE Conference on Computer Vision and Pattern
  Recognition (CVPR)}, 2016.

\bibitem{deng2009imagenet}
Jia Deng, Wei Dong, Richard Socher, Li-Jia Li, Kai Li, and Li Fei-Fei.
\newblock Imagenet: A large-scale hierarchical image database.
\newblock In {\em 2009 IEEE conference on computer vision and pattern
  recognition}, pages 248--255. Ieee, 2009.

\bibitem{goyal2017accurate}
Priya Goyal, Piotr Doll{\'a}r, Ross Girshick, Pieter Noordhuis, Lukasz
  Wesolowski, Aapo Kyrola, Andrew Tulloch, Yangqing Jia, and Kaiming He.
\newblock Accurate, large minibatch sgd: Training imagenet in 1 hour.
\newblock {\em arXiv preprint arXiv:1706.02677}, 2017.

\bibitem{he2016deep}
Kaiming He, Xiangyu Zhang, Shaoqing Ren, and Jian Sun.
\newblock Deep residual learning for image recognition.
\newblock In {\em Proceedings of the IEEE conference on computer vision and
  pattern recognition}, pages 770--778, 2016.

\bibitem{huang2017adain}
Xun Huang and Serge Belongie.
\newblock Arbitrary style transfer in real-time with adaptive instance
  normalization.
\newblock In {\em ICCV}, 2017.

\bibitem{ioffe2015batch}
Sergey Ioffe and Christian Szegedy.
\newblock Batch normalization: Accelerating deep network training by reducing
  internal covariate shift.
\newblock In {\em International Conference on Machine Learning}, pages
  448--456, 2015.

\bibitem{krizhevsky2012imagenet}
Alex Krizhevsky, Ilya Sutskever, and Geoffrey~E Hinton.
\newblock Imagenet classification with deep convolutional neural networks.
\newblock In {\em Advances in neural information processing systems}, pages
  1097--1105, 2012.

\bibitem{li2016revisiting}
Yanghao Li, Naiyan Wang, Jianping Shi, Jiaying Liu, and Xiaodi Hou.
\newblock Revisiting batch normalization for practical domain adaptation.
\newblock {\em arXiv preprint arXiv:1603.04779}, 2016.

\bibitem{lin2014microsoft}
Tsung-Yi Lin, Michael Maire, Serge Belongie, James Hays, Pietro Perona, Deva
  Ramanan, Piotr Doll{\'a}r, and C~Lawrence Zitnick.
\newblock Microsoft coco: Common objects in context.
\newblock In {\em European conference on computer vision}, pages 740--755.
  Springer, 2014.

\bibitem{luo2018differentiable}
Ping Luo, Jiamin Ren, Zhanglin Peng, Ruimao Zhang, and Jingyu Li.
\newblock Differentiable learning-to-normalize via switchable normalization.
\newblock In {\em International Conference on Learning Representations}, 2019.

\bibitem{milletari2016v}
Fausto Milletari, Nassir Navab, and Seyed-Ahmad Ahmadi.
\newblock V-net: Fully convolutional neural networks for volumetric medical
  image segmentation.
\newblock In {\em 2016 Fourth International Conference on 3D Vision (3DV)},
  pages 565--571. IEEE, 2016.

\bibitem{pan2018IBN-Net}
Xingang Pan, Ping Luo, Jianping Shi, and Xiaoou Tang.
\newblock Two at once: Enhancing learning and generalization capacities via
  ibn-net.
\newblock In {\em ECCV}, 2018.

\bibitem{sutskever2013Momentum}
Ilya Sutskever, James Martens, George Dahl, and Geoffrey Hinton.
\newblock On the importance of initialization and momentum in deep learning.
\newblock In {\em International conference on machine learning}, pages
  1139--1147, 2013.

\bibitem{ulyanov2016instance}
Dmitry Ulyanov, Andrea Vedaldi, and Victor Lempitsky.
\newblock Instance normalization: The missing ingredient for fast stylization.
\newblock {\em arXiv preprint arXiv:1607.08022}, 2016.

\bibitem{zhou2016semantic}
Bolei Zhou, Hang Zhao, Xavier Puig, Sanja Fidler, Adela Barriuso, and Antonio
  Torralba.
\newblock Semantic understanding of scenes through the ade20k dataset.
\newblock {\em arXiv preprint arXiv:1608.05442}, 2016.

\bibitem{zhou2017scene}
Bolei Zhou, Hang Zhao, Xavier Puig, Sanja Fidler, Adela Barriuso, and Antonio
  Torralba.
\newblock Scene parsing through ade20k dataset.
\newblock In {\em Proceedings of the IEEE Conference on Computer Vision and
  Pattern Recognition}, 2017.

\bibitem{CycleGAN2017}
Jun-Yan Zhu, Taesung Park, Phillip Isola, and Alexei~A Efros.
\newblock Unpaired image-to-image translation using cycle-consistent
  adversarial networks.
\newblock In {\em Computer Vision (ICCV), 2017 IEEE International Conference
  on}, 2017.

\end{thebibliography}
}

\end{document}